\documentclass[review]{elsarticle}

\usepackage{lineno,hyperref}

\journal{Journal of \LaTeX\ Templates}
\usepackage{amsmath}
\usepackage{gensymb}
\usepackage{multirow}
\usepackage{booktabs}

\begin{document}

\begin{frontmatter}

\title{JointsGait : Gait Recognition based on Graph Convolutional Networks and Joints Relationship Pyramid Mapping}


\author[mymainaddress]{Na Li}
\ead{cvlina@mail.nwpu.edu.cn}

\author[mymainaddress]{Xinbo Zhao\corref{mycorrespondingauthor}}
\cortext[mycorrespondingauthor]{Corresponding author}
\ead{xbozhao@nwpu.edu.cn}

\author[mysecondaryaddress]{Chong Ma}
\ead{mc-npu@mail.nwpu.edu.cn}

\address[mymainaddress]{National Engineering Laboratory for Integrated Aero-Space-Ground-Ocean Big Data Application Technology, School of Computer Science, Northwestern Polytechnical University, China}
\address[mysecondaryaddress]{School of Automation, Northwestern Polytechnical University, China}





\begin{abstract}
Gait, as one of unique biometric features, has the advantage of being recognized from a long distance away, can be widely used in public security. Considering 3D pose estimation is more challenging than 2D pose estimation in practice , we research on using 2D joints to recognize gait in this paper, and a new model-based gait recognition method JointsGait is put forward to extract gait information from 2D human body joints. Appearance-based gait recognition algorithms are prevalent before. However, appearance features suffer from external factors which can cause drastic appearance variations, e.g. clothing. Unlike previous approaches, JointsGait firstly extracted spatio-temporal features from 2D joints using gait graph convolutional networks, which are less interfered by external factors. Secondly, Joints Relationship Pyramid Mapping (JRPM) are proposed to map spatio-temporal gait features into a discriminative feature space with biological advantages according to the relationship of human joints when people are walking at various scales. Finally, we design a fusion loss strategy to help the joints features to be insensitive to cross-view. Our method is evaluated on two large datasets, Kinect Gait Biometry Dataset and CASIA-B. On Kinect Gait Biometry Dataset database, JointsGait only uses corresponding 2D coordinates of joints, but achieves satisfactory recognition accuracy compared with those model-based algorithms using 3D joints. On CASIA-B database, the proposed method greatly outperforms advanced model-based methods in all walking conditions, even performs superior to state-of-art appearance-based methods when clothing  seriously affect people’s appearance. The experimental results demonstrate that JointsGait achieves the state-of-art performance despite the low dimensional feature (2D body joints) and is less affected by the view variations and clothing variation.
\end{abstract}


\end{frontmatter}


\section{Introduction}

Gait recognition contributes to discriminate individuals by their walking way. Contrast to other biometric features, gait is well suitable for long-distance human identification. Therefore, there are bright prospects for gait recognition in numerous applications such as visual surveillance, forensic identification, and national security.

Unfortunately, gait recognition is easily affected by several factors such as clothing, view and carrying. Some earlier gait recognition methods try to extract robust gait feature by modeling human body and capturing motion patterns of walking \cite{nixon1999automatic,tanawongsuwan2001gait,wang2004fusion}. Using the body motion to recognize gait is straightforward and reasonable, however, accurately locating and tracking human body parts is a tough challenge. 

\begin{figure}[htb]
 
\centering
\includegraphics[scale=0.3]{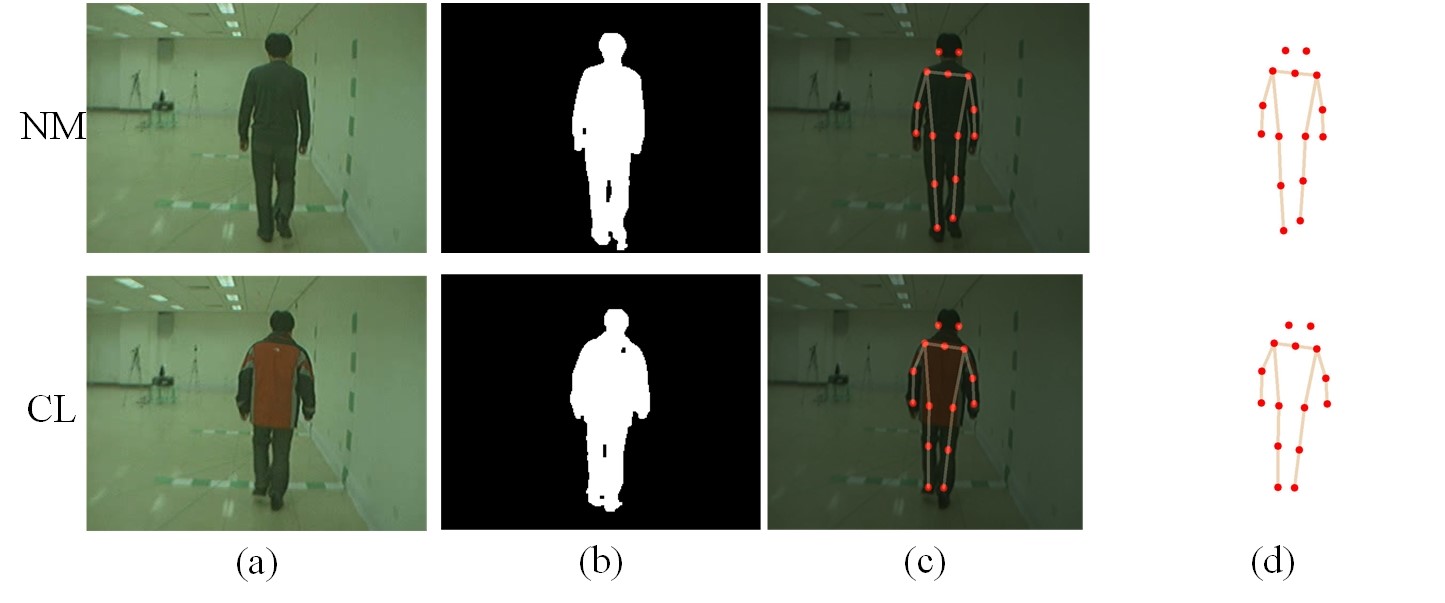}
\caption{ The samples of two common walking conditions. NM is normal condition, CL is the man wearing a thick coat. (a) is original video frames, (b) is silhouettes, (c) is human pose estimation results, (d) is human skeletons.}
\label{fig1}
\end{figure}

Hence, unlike model-based gait recognition methods, those methods using appearance features \cite{han2005individual,chao2019gaitset} are prevailing gait recognition algorithms in the past two decades, whose raw input data usually are human silhouettes (e.g. Figure \ref{fig1} (b)) obtained from background subtraction. These methods can achieve well performance without interference factors. However, when human shape changes greatly in practice (e.g. Figure \ref{fig1} CL), the appearance-based methods’ performance may decrease severely. In addition, the recognition accuracy of appearance-based algorithms counts heavily on the clarity of the silhouettes. If the camera has a certain movement, it is hard to obtain reliable silhouettes and satisfactory gait recognition results. Instead, model-based features are relatively robust to human shape and appearance, because they are extracted from human body structure and movements (e.g. Figure \ref{fig1} (d)). 

Significant progress has been made in pose estimation \cite{cao2017realtime} in recent years, which brings more hope to the model-based algorithms. Considering 3D pose estimation is more challenging than 2D pose estimation since it needs to predict the depth information of body joints, we research on using 2D joints to recognize gait in this paper. A new model-based gait recognition algorithm, JointsGait, is proposed based on gait GCNs, which exploits high-performance gait features from human body structure to handle many variations. Firstly, we extract spatio-temporal gait features by constructing spatiotemporal joints graph from video sequence. Secondly, we propose Joints Relationship Pyramid Mapping (JRPM) to obtain the final feature representation, which maps spatio-temporal gait features into a more discriminative space according to human body structure and walking habit. Finally, we research on a fusion loss strategy to help the joint features be insensitive to view variations. Experimental results demonstrate that JointsGait achieves the state-of-art performance and is robust to view and clothing variations. 

The main contributions of this paper are summarized as follows:
\begin{itemize}
\item We present JointsGait, a novel model-based gait recognition method that obtains state-of-the-art performance despite the low dimensional feature (2D body joints) based on graph convolutional networks.  
\item According to the relationship of human joints when people are walking, we propose a biologically reasonable pyramid pooling method JRPM, which is greatly helpful to improve the recognition accuracy in all walking conditions by enhancing local features.
\item To solve the problem of cross-view, that is to reduce the impact of viewpoints when recognizing gait, we design a fusion loss function.

\end{itemize}

\section{Related work}

Here, we briefly review appearance-based gait recognition algorithms, model-based gait recognition algorithms, graph convolutional neural networks and pyramid pooling.

\subsection{Appearance-based gait recognition algorithms}
Appearance-based gait recognition algorithms usually extract different gait information from the human silhouettes. By rendering pixel level operators on silhouettes, some methods contribute to generate different gait template, e.g. Gait energy image (GEI)\cite{han2005individual}, Chrono-Gait Image (CGI)\cite{wang2011human}. Among template matching approaches, the projection between different views are learned in View Transformation Model(VTM)\cite{makihara2006gait} . Gait data with arbitrary views are transformed to a specific angle in Stacked Progressive Auto-Encoders (SPAE) \cite{yu2017invariant}. View-invariant Discriminative Projection (ViDP) \cite{hu2013view} makes the templates into a latent space for learning a view-invariance representation. Instead of using gait templates, some methods \cite{chao2019gaitset,yu2017gaitgan,wu2016comprehensive,he2018multi,yu2019gaitganv2} directly utilize silhouettes with the help of deep learning algorithms. \cite{wu2016comprehensive} is the first one applying a deep learning method to recognize gait from human silhouettes. Besides, GaitSet \cite{chao2019gaitset}, which regards gait as an independent silhouette set rather than continuous silhouettes \cite{wu2016comprehensive} , surpasses previous state-of-the-art approaches’ the accuracies of gait recognition.

Although these algorithms perform well in cross-view condition, they are easily affected by some variations because they are largely depending on human appearance and shape. Besides, their input, silhouettes, are also difficult to obtain when the camera moving.

\subsection{Model-based gait recognition algorithms}
Model-based algorithms obtain gait information through modeling body structure and different body parts’ motion patterns. Unlike appearance-based algorithms, if human body structure is highly correctly modeled, model-based algorithms could be robust to many variations. However, it is a challenging task. Therefore, model-based algorithms are not popular in the past two decades. Manually marking body parts even applied in some earlier model-based algorithms. \cite{nixon1999automatic} simulates legs by utilizing a simple stick model and then simulate the leg movement by using an articulate pendulum. Wang et al. \cite{wang2004fusion} assert that each joint’s angle changes in temporal domain is beneficial for gait recognition.

Recently, some methods recognize gait by skeleton and joints. Using skeletal data obtained the low-cost Kinect sensor , \cite{kastaniotis2016pose,andersson2015person,yang2016relative,li2017dynamic,khamsemanan2017human,sun2018view,liu20193d} propose various gait recognition methods. The methods of \cite{andersson2015person,yang2016relative,khamsemanan2017human,sun2018view} extract gait features by manually defined gait features. Li et al. \cite{khamsemanan2017human} use a dynamic LSTM network to learn descriptive sequence characteristics directly from the raw skeleton data. Liu et al. \cite{liu20193d} design a CNN-LSTM network to extract the temporal-spatial deep feature information from Skeleton Gait Energy Image. These methods demonstrate that the joints contain adequate information for human identification.

Considering Kinect sensors are not commonly used in video surveillance than RGB cameras, \cite{liao2017pose,an2018improving,liao2020model} use body joints directly estimated from original video by pose estimation methods to recognize gait. Although  \cite{liao2017pose,an2018improving,liao2020model}  are different from the previous methods of marking human body parts by manual methods or special equipment, they still manually define gait features. Besides, to deal with view changes, \cite{an2018improving,liao2020model} converts 2D joint points into 3D joint points, which increase computational cost. However, our method exploits reliable gait features using only 2D joints with considerably less effort in manual design and has excellent performance.
\subsection{Graph convolutional neural networks}
Nowadays, with advancement of deep neural networks, graph convolutional networks (GCNs) has developed rapidly \cite{defferrard2016convolutional,2015Gated,hamilton2017inductive,monti2017geometric,kipf2018neural}. Using arbitrarily structured graphs, GCNs extend convolutional neural networks (CNNs) into high-dimensional domains, which can be classified into spectral perspective methods and spatial perspective methods. The former applies CNNs to the spectral domain and transform graph into a spectrum \cite{defferrard2016convolutional,2015Gated}. While spatial perspective methods define parameterized filters by using graph convolutions \cite{hamilton2017inductive,monti2017geometric,kipf2018neural}, which is similar to the convolution operation on images. This paper follows the spatial perspective method.
\subsection{Pyramid Pooling}
To gather local and global information, many methods use pyramid pooling rather than simple global pooling to make deep networks focusing on different size feature in many computer vision tasks. By pooling local spatial feature bins, spatial Pyramid Pooling network (SPP) \cite{he2015spatial} preserves spatial information. \cite{zhao2017pyramid} also uses similarly pyramid pooling module, which slices the feature map into different sub-regions by pyramid level pooling and forms different locations’ pooled representation. In human re-identification task, Horizontal Pyramid Pooling (HPP) \cite{fu2019horizontal} horizontally separate features into many spatial bins by multiple scales.

Considering advantages of pyramid pooling and the physiological connection between human body joints, this paper divides the spatio-temporal gait features into different body sub-regions according to human body structure and walking habit.

\section{JointsGait}
\begin{figure}[htb]
\centering
\includegraphics[scale=0.23]{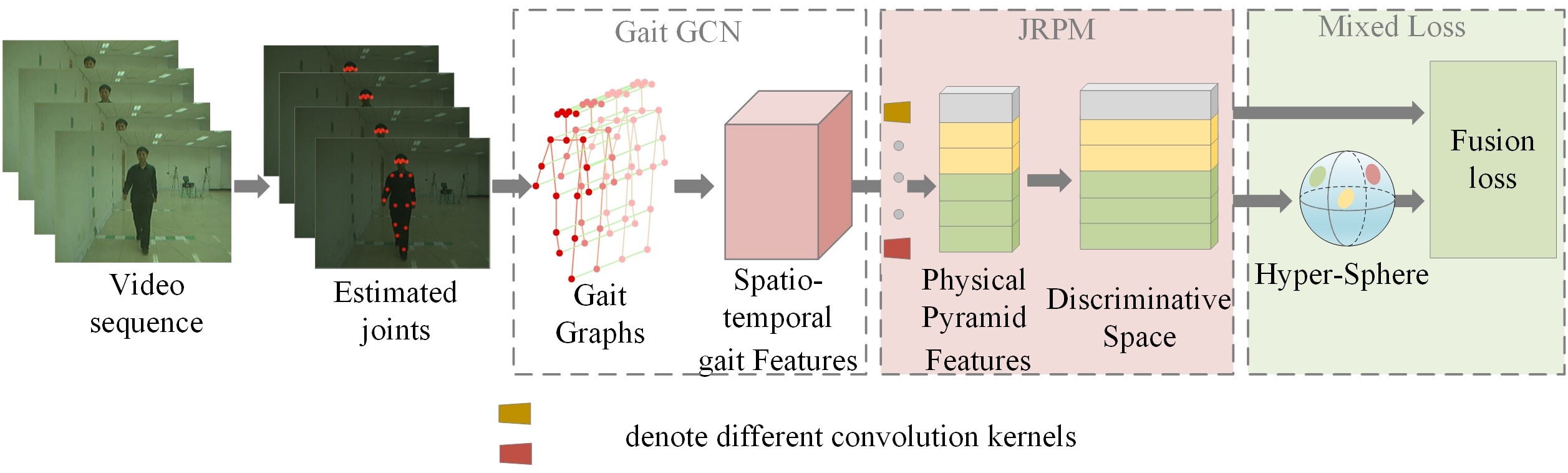}
\caption{ The framework of the proposed JointsGait. Human body joints estimated from video sequence is used to build gait graph. Through gait graph convolutional networks, spatio-temporal gait features are extracted, which is mapping into discriminative space by JRPM. A fusion loss strategy is applied to optimize our method.}
\label{fig2}
\end{figure}

We introduce JointsGait in this section, which learns discriminative gait information from the human body structure and movements instead of human shape and human appearance, and the overall pipeline is shown in Figure \ref{fig2}. The proposed method takes estimated human body joints as input, next constructs gait graph convolutional networks to extract spatio-temporal gait features, then uses our JRPM to map spatio-temporal gait features into a more discriminative space, eventually handles variations better with the help of a fusion loss strategy. The following parts of this section present implementation details.

\subsection{Graph convolutional neural networks}
\subsubsection{Gait Graph}
\begin{figure}[htb]
 \centering
\includegraphics[scale=0.9]{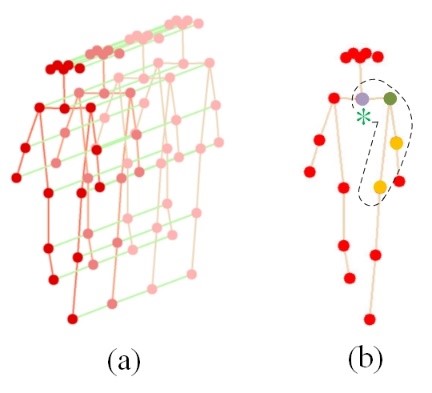}
\caption{(a) An example of the spatio-temporal gait graph. (b) An example of the mapping strategy, and different colors represent different subsets.}
\label{fig3}
\end{figure}

For modeling the structured information among human body joints when people are walking, a spatio-temporal gait graph is made by following the work of ST-GCN \cite{yan2018spatial}. An example of the constructed gait graph is shown in Figure \ref{fig3} (a), where dots represent the joints, red lines edges stand for joints’ physical connections in spatial dimension, and the green lines represent temporal edges, which connect two neighboring frames’ the corresponding joints in temporal dimension. These joints are estimated by openpose\cite{cao2017realtime}, which uses 18 joints to represent a person body. And the coordinate vector of each joint is defined to its attribute.

\subsubsection{Graph convolution}
In ST-GCN \cite{yan2018spatial}, the graph convolution is applied on vertex ${{v}_{i}}$  in the spatial dimension as Eq.\ref{eq1} shows.

\begin{equation}
{{f}_{out}}({{v}_{i}})=\sum\limits_{{{v}_{j}}\in {{B}_{i}}}{\frac{1}{{{Z}_{ij}}}{{f}_{in}}({{v}_{j}})}\cdot w({{l}_{i}}({{v}_{j}}))
\label{eq1}
\end{equation}

where ${{f}_{in}}$ is input feature and ${{f}_{out}}$ is output feature. ${{B }_{i}}$ is the sampling region of the convolution for ${{v}_{i}}$, which defined as the target vertex ${{v}_{i}}$’s 1-distance neighbors. And the number of vertexes in ${{B }_{i}}$ is variable. $w$ is the weight function that offers a weight vector and calculates the inner product with input gait feature. ${{l}_{i}}$ is a mapping function is designed to maps each vertex with a special weight vector. This strategy is illustrated in Figure \ref{fig3} (b), where $*$ stands for the skeleton’s center of gravity and the area surrounded by the dotted line is ${{B }_{i}}$. There are three subsets of ${{B }_{i}}$ are naturally divided by the strategy : the vertex itself is ${{S}_{i1}}$ (the green circle in Figure \ref{fig3} (b)); the centripetal subset is ${{S}_{i2}}$ (the purple circle in Figure \ref{fig3} (b)), which includes the adjacent vertexes that are next to the gravity center; ${{S}_{i3}}$ is otherwise centrifugal subset (the orange circle in Figure \ref{fig3} (b)). ${{Z}_{ij}}$ balances the contribution of these subsets, which is used as the normalizing term.
\subsubsection{Implementation}
For obtaining the spatio-temporal gait features, multiple layers of spatio-temporal graph convolution operations are applied on gait graph. 

However, it is not straightforward to implement graph convolution in the spatial dimension. Concretely, the feature map of  JointsGait is a $N\times C\times T$ tensor, where $N$ represents the number of vertexes in video sequence, $C$denotes the number of channels and the temporal length is $T$. To implement the graph convolution, Eq.\ref{eq1} is transformed into Eq.\ref{eq2}:

\begin{equation}
{{\mathbf{f}}_{out}}=\sum\limits_{k}^{{{K}_{v}}}{{{\mathbf{W}}_{k}}({{\mathbf{f}}_{in}}\mathbf{\Lambda }_{k}^{-\tfrac{1}{2}}{{\mathbf{A }}_{k}}\mathbf{\Lambda }_{k}^{-\tfrac{1}{2}})}\odot {{\mathbf{M}}_{k}} 
\label{eq2}
\end{equation}

where ${{K}_{v}}$ is the kernel size of the spatial dimension, which is set to 3 with the partition strategy mentioned above. ${{\mathbf{A}}_{k}}$ is the $N\times N$ adjacency matrix, whose element ${{\mathbf{A}_{k}^{ij}}}$ shows whether the vertex ${{v}_{j}}$ belongs to the subset ${{S}_{ik}}$ of vertex ${{v}_{i}}$. The normalized diagonal matrix  $\Lambda _{k}^{ii}=\sum\nolimits_{j}{(\mathbf{A}_{k}^{ij})}+\alpha $, where $\alpha \text{=}0.001$ for avoiding empty rows in ${{\mathbf{A}}_{k}}$. ${{\mathbf{W}}_{k}}$ is the weight matrix, which stands for the weighting function  $w$ in Eq.\ref{eq1}. ${{\mathbf{M}}_{k}}$ is a $N\times N$ attention map which shows the importance of each vertex. $\odot $ is the dot product.

The implementation of the graph convolution in the temporal dimension is similar to the classical convolution operation, because the number of neighbors for each vertex can be fixed in temporal dimension (corresponding joints in consecutive frames). Hence, we apply a convolution kernel ${{K}_{t}}$ on the spatial gait feature ${{\mathbf{f}}_{out}}$ in temporal dimension to obtain spatio-temporal gait features ${{F}_{ST}}$: 

\begin{equation}
{{F}_{ST}}={{\mathbf{f}}_{out}}*{{K}_{t}}  
\label{eq3}
\end{equation}

\subsection{Joints Relationship Pyramid Mapping}
JRPM is designed to map spatio-temporal gait features into a more discriminative space through enhancing the discriminative gait feature of partial human body, which is consist of Joints Relationship Pyramid Pooling (JRPP) and separate fully connect layers (FC).
 
JRPP is inspired by Horizontal Pyramid Pooling (HPP)\cite{fu2019horizontal}. HPP horizontally divided the feature maps into multiple scrips. And the appearance-based model Gaitset \cite{chao2019gaitset} uses HPP to obtain well-performance local features. However, HPP is not suitable for skeleton-based method, because gait graph constructd by joints are not Euclidean structure. Hence, we propose JRPP to extract multiple scales of local gait features according to the relationship of human joints when people are walking. 

Specifically, JRPP has $S$ scales. On scale $s\in 1,2,...,S$, the spatio-temporal gait features ${{F}_{ST}}$ is split into ${{G}_{s}}$ groups local features, and there are $\sum\limits_{s=1}^{S}{\sum\limits_{g=1}^{{{G}_{S}}}{{{F}_{s,g}}}}$ in total, where ${{F}_{s,g}}$ is a local feature, $s$ denotes the index of scale and $g$ denotes the index of local feature in each scale. For instance, ${{F}_{3,1}}$ means the first local feature in third scale. Then, each local feature ${{F}_{s,g}}$ is pooled by convolution kernel ${{k}_{s,g}}$ to generate physical pyramid feature ${{F}_{PP}}$:

\begin{equation}
{{F}_{PP}}=\sum\limits_{s=1}^{S}{\sum\limits_{g=1}^{{{G}_{S}}}{{{F}_{s,g}}}}*{{k}_{s,g}}                       
\label{eq4}
\end{equation}

where the size of ${{F}_{s,g}}$ is $N\times C\times {{J}_{s,g}}\times T$, ${{J}_{s,g}}$ is the number of joints in local body and the size of ${{k}_{s,g}}$ is ${{J}_{i,j}}\times T$. 

\begin{figure}[htb]
\centering
\includegraphics[scale=0.3]{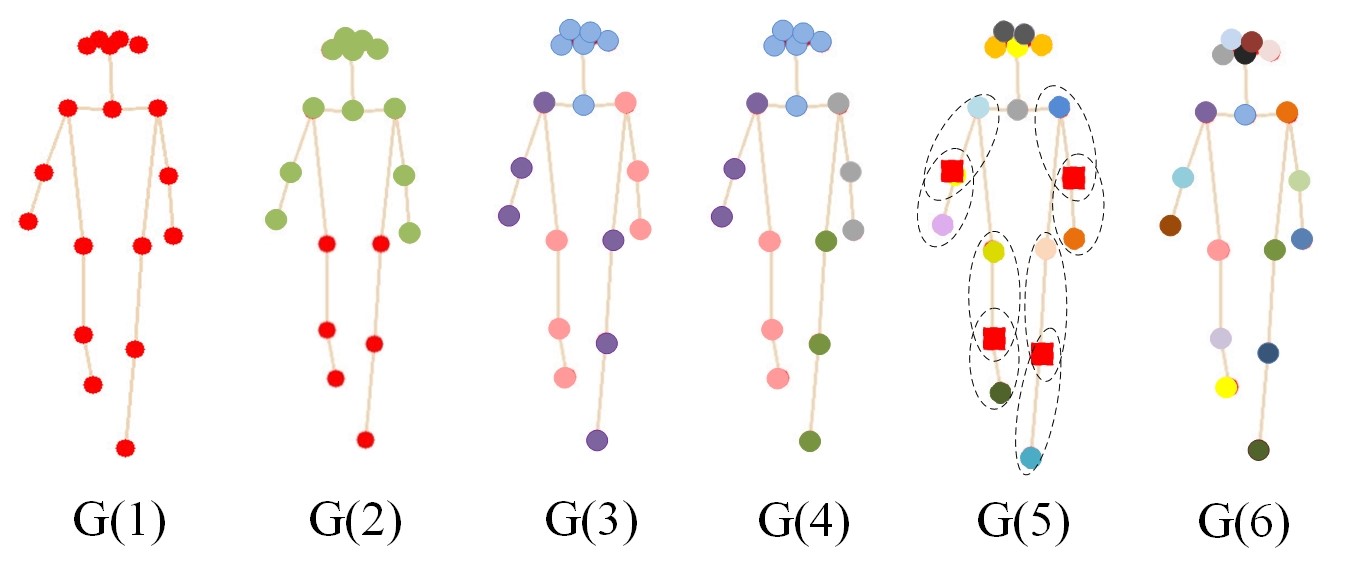}
\caption{The ways of grouping joints on different scales.}
\label{fig4}
\end{figure}

According to different prior knowledge, the body joints are divided into different groups on different scale until each joint is divided into a group, forming 6-layers gait feature pyramid. For example, Figure \ref{fig4}  G($i$) intuitively show ways of grouping joints estimated by   openpose\cite{cao2017realtime} on  scale $i$. In G($i$), the joints with same color or in a same ellipse are divided into a group, and the red square joints are shared by two groups in G(5). Specifically, the whole body in a group in G(1); The upper and lower body are divided into two groups in G(2) according to horizontal plane in anatomy ; In G(3), left arm and right leg are designed in a group according to human general walking habit, and same for right arm and left leg, because when people walk, they always naturally swing their arms, which is exactly opposite to the direction of their legs; In G(4), the limbs were divided into four groups, and head is a group; In G(5), the limbs are divided into upper and lower parts and there are 12 groups in total; Finally, 18 joints are divided into 18 different groups in G(6). Hence, the number of groups of local features ${{G}_{s}}\in \left[ 1,2,3,5,12,18 \right]$ on scale $s\in \left\{ 1,2,...,6 \right\}$.

\begin{figure}[htb]
\centering
\includegraphics[scale=0.7]{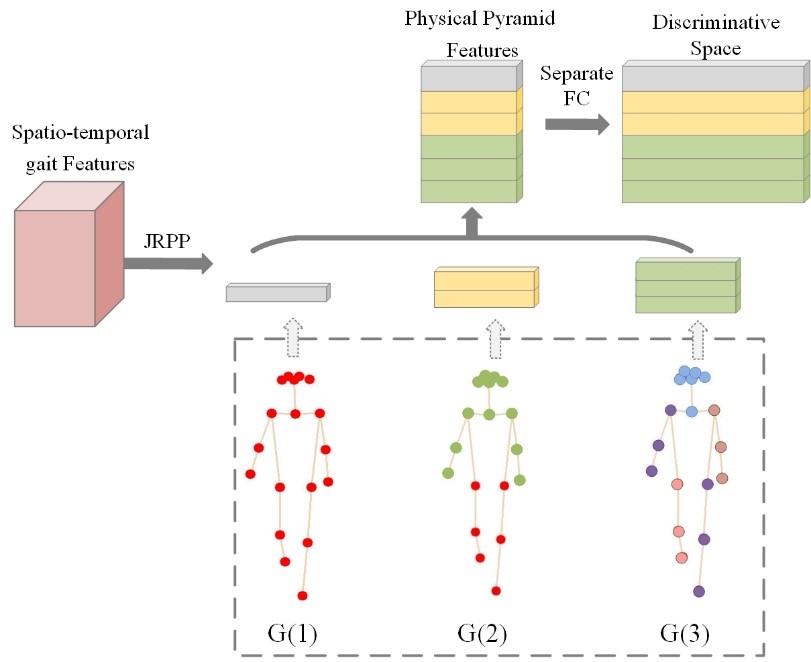}
\caption{Overview of the proposed JRPM approach. The JRPP is applied on spatio-temporal gait features to produce physical pyramid features on 3 scales. Finally, these physical pyramid features is mapped into the discriminative space by separating fully connect layers. }
\label{fig5}
\end{figure}

 In section 4.2, we apply JRPP with different pyramid scales on ${{F}_{ST}}$, and experimental results show that JRPM obtains satisfactory performance on 3 pyramid scales. Hence, as shown in Figure \ref{fig5}, we adopt 3 pyramid scales within JRPP. 

Independent fully connect layers are used to map the pooled physical pyramid features ${{F}_{PP}}$ into the discriminative space to obtain the final gait feature representation. In such a way, the discernment of human gait could be obtained from coarse to fine.

\subsection{Fusion loss Strategy}
\begin{figure}[htb]
\centering
\includegraphics[scale=0.8]{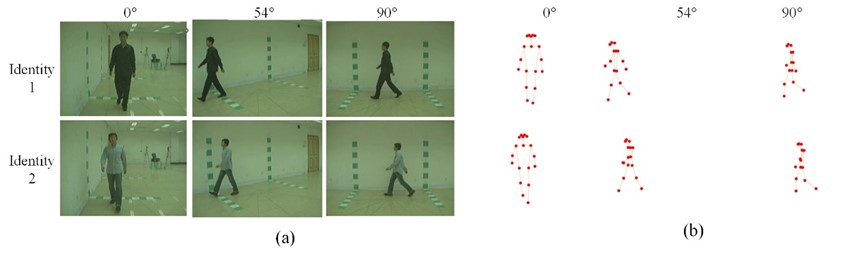}
\caption{The samples of view variation and the view angles are \{0 \degree, 54 \degree, 90 \degree \}. (a) is original video frames (b) is corresponding skeleton data of (a).}
\label{fig6}
\end{figure}

The extracted gait features contain both spatiotemporal information and physiological characteristics. However, due to the influence of views, different views of the same identity usually have huge visual differences, besides it is even possible that different identities in same view closely resembled each other, while the same identity in the different view seem to be quite different, especially in skeleton data(e.g. Figure \ref{fig6} (b)). To solve the problem of cross-view, that is to reduce the impact of viewpoints when recognizing gait, we design a fusion loss function.

We regard different viewpoints of a person as different clusters of a class, so cross-view gait recognition problem could be abstracted as a clustering problem. Therefore, we consider two aspects for designing the loss function: one is that the loss function should reduce the intra-class distance, meanwhile, increase inter-class distance. Another is that it may be difficult to find the cluster center, because the difference of clusters of same class may be very large.

Among existing loss functions, the triplet loss \cite{schroff2015facenet} is defined as Eq.\ref{eq5} shows that the distance between anchor point ${{x}_{a}}$ and a positive point ${{x}_{p}}$ belonging to the same class  ${{y}_{a}}$ is less than the distance between anchor point ${{x}_{a}}$ and a negative point ${{x}_{n}}$ belonging to another class ${{y}_{n}}$ more than a margin $m$. The advantage of triplet loss is that although all points of the same class will eventually form a cluster, they do not need to collapse to a point , and only need to be closer to each other , which is consistent with our task.

\begin{equation}
{{L}_{tri}}=\sum\limits_{\begin{smallmatrix} 
 a,p,n \\ 	
 {{y}_{a}}={{y}_{p}}\ne {{y}_{n}} 
\end{smallmatrix}}{{{\left[ m+{{D}_{a,p}}-{{D}_{a,n}} \right]}_{+}}}                       
\label{eq5}
\end{equation}

Therefore, we adopt triplet loss to optimize our network at first, and advised in \cite{hermans2017defense}, we use improved triplet loss based on batch to accelerate network convergence. The size of a batch is $PK$, which is constituted by $P$ classes (identities), and $K$ images of each class (views) based on stochastic sampling. And we regard different identity at same views as hard negatives, and same identity at different views as hard positives. The triplets are formed by the hardest positive and the hardest negative samples within the batch, which is used to compute triplet loss as shown in Eq. \ref{eq6}.

\begin{equation}
{{L}_{tri\_BH}}=\overbrace{\sum\limits_{i=1}^{P}{\sum\limits_{a=1}^{K}{{}}}}^{all\text{ }anchors}{{[m+\overbrace{\underset{p=1\cdots K}{\mathop{\max }}\,D(x_{a}^{i},x_{p}^{i})}^{hardest\text{ }positive}-\underbrace{\underset{\begin{smallmatrix} 
 j=1\cdots P \\ 
 n=1\cdots K 
 \\ 
 j\ne i 
\end{smallmatrix}}{\mathop{\min }}\,D(x_{a}^{i},x_{n}^{j})}_{hardest\text{ }negative}]}_{+}} 
\label{eq6}
\end{equation}

where $x_{j}^{i}$ denotes to the $j\text{-th}$ feature belongs to the $i\text{-th}$ identity. 

However, in the real experiment, as shown in section 4.2, we found that although triplet loss can widen inter-class distance and reduce intra-class distance to a certain extent, but the cross-view problem is still difficult for it. Thus, we need another loss function to increase inter-class distance. Arcface loss\cite{deng2019arcface}, as improvement of softmax loss, is proposed to solve face recognition. By mapping features to a hypersphere, inter-class distance can be increased. And its calculation formula is as follows:
\begin{equation}
{{L}_{arc}}=-\frac{1}{N}\sum\limits_{i=1}^{N}{\log \frac{{{e}^{c(\cos ({{\theta }_{{{y}_{i}}}}+m))}}}{{{e}^{c(\cos ({{\theta }_{{{y}_{i}}}}+m))}}+\sum{_{j=1,j\ne {{y}_{i}}}^{n}{{e}^{c\cos {{\theta }_{j}}}}}}}
\label{eq7}
\end{equation}

Where the angle between the ground truth weight ${{W}_{y}}$ and the feature and is denoted as $\theta $, $m$ is an angular margin penalty, and $c$ is the feature scale.

But arcface loss may also increase intra-class distance because the skeleton data of the same person varies greatly from different viewpoints. Therefore, the fusion of triplet loss and arcface loss needs to be weighed to deal with the problem of cross-view. The triplet loss should play a major role, and the arcface loss is also indispensable. Finally, we define the loss function of skeleton-based gait recognition as Eq. \ref{eq8}, and $\lambda $ is set to 0.9 in our experiments, which is the best value in our experiments in section 4.2.  
\begin{equation}
L=\lambda {{L}_{tri\_BH}}+(1-\lambda ){{L}_{arc}}
\label{eq8}
\end{equation}

\section{Experimental results and analysis}
Our empirical experiments mainly consist of four parts. The first part is ablation experiments conducted on most popular gait database CASIA-B database \cite{yu2006framework} because this database contains multiple covariates. In the second part, we compare JointsGait with state-of-the-art model-based methods \cite{andersson2015person,yang2016relative,li2017dynamic,khamsemanan2017human,sun2018view,liu20193d} on a large skeleton gait dataset Kinect Gait Biometry Dataset \cite{andersson2015person}. Although Kinect Gait Biometry Dataset provides 3D body joints by a Kinect sensor, JointsGait only uses the corresponding 2D coordinates of joints to evaluate its ability to recognize gait through 2D joints. Because RGB cameras are commonly used in video surveillance, not Kinect sensors, we focus on evaluating the performance of JointsGait on popular CASIA-B database captured by RGB cameras. Therefore, in third part, we evaluate the robustness of JointsGait for cross-view and walking condition variations,and compare it with advanced model-based methods \cite{liao2017pose,an2018improving,liao2020model} on CASIA-B. Finally, JointsGait is compared with state-of-art appearance-based gait recognition methods\cite{chao2019gaitset,yu2017invariant,yu2017gaitgan,he2018multi,yu2019gaitganv2,liu2020gait,zhang2020learning} when clothing variation seriously affect people’s appearance.

\subsection{Datasets and Experimental settings}

When evaluating gait recognition method JointsGait, we need gait databases that can provide body joints coordinates or contain original video due to the human pose only could be estimated from original video and rather than silhouettes. Kinect Gait Biometry Dataset can provide 3D coordinates of 20 body joints by a Kinect V1 sensor. Unfortunately, most of public gait databases captured by color cameras don’t releasee original video due to the privacy issue. CASIA-B is adopted in our experiments not only because it contains 11 view variations along with clothing and carrying covariate conditions, but also because it provides the original video.

\textbf{CASIA-B} As one of popular public gait datasets, CASIA-B dataset is widely applied in gait recognition research, which contains the gait data of 124 subjects (31 females and 93 males). The gait data of each subject is consist of 6 clips in normal walking (NM) condition, 2 clips in bagged walking (BG) condition, and  2 clips in coat walking (CL) condition, and there are 10 clips in total .And 11 cameras are used to simultaneously capture data from 11 views, the view angles are \{0\degree, 18\degree, 36\degree, 54\degree, …, 180\degree\}. 

In all the experiments on CASIA-B, we use the public available pose estimation method OpenPose \cite{cao2017realtime} to obtain 18 human joints on every frame of the clips. The mini-batch consists of the manner introduced in section 3.3 with $P\text{=8}$ and $K=16$. We randomly select 120 frames per clips. There are 9 spatial temporal graph convolution layers in the gait GCNs model. The output of first three layers is 64 dimensions, the output of 4 to 6 layers is 128 dimensions, besides, the output of last three layers have 256 dimensions. The number of scales $S=3$ in JRPM. JRPP pools 256 dimensions spatio-temporal gait features into $256\times 21$dimensions physical pyramid feature, then maps $512\times 21$dimensions discriminative space. The margin of triplet loss ${{L}_{tri\_BH}}$ is set to 0.2 and the margin of arcface loss ${{L}_{arc}}$ is set to 0.35. We trained our model on CASIA-B for about 80K iterations with 2 NVIDIA 1080TI GPUs.

\textbf{Kinect Gait Biometry Dataset}  This skeleton gait dataset contains walking sequences from 164 subjects (the initial version is 140 subjects), and each has four or five sequences. In the data capturing procedure, each subject was asked to walk in front of the Kinect in a semi-circular trajectory. A spinning dish helps Kinect to keep the subject always in the center of its view.

Following the protocol of \cite{li2017dynamic}, we randomly select 140 subjects of Kinect Gait Biometry Dataset to conduct experiments by using the 10-fold cross validation technique. We trained our model for about 2K iterations on this dataset with 1 NVIDIA 1080TI GPU. When JRPM is applied on Kinect Gait Biometry Dataset , the pooling of the same body part is the same as CASIA-B database regardless of how many joint points are marked. And other experimental settings are the same as those on CASIA-B database.

\subsection{Ablation Experiments on CASIA-B database}
 For demonstrating the effectiveness of each part and settings of JointsGait, we design several ablation experiments with different settings on CASIA-B database as shown in Table \ref{table1}, including different loss function strategies, different pyramid scales of JRPM. And all unrelated settings in these studies are the same. Except the identical-view cases, results are accuracies averaged on rest 10 views.

Since our spatio-temporal gait graph is made by following the work of ST-GCN \cite{yan2018spatial}, ST-GCN \cite{yan2018spatial} is also tested on CASIA-B database as baseline. The experimental results are shown in the first row of Table  \ref{table1}. Its performance is very poor, because ST-GCN \cite{yan2018spatial} is designed for action classification and is not good at cross-view problem. However, the spatio-temporal features extracted from GCNs are useful for gait recognition.

\textbf{Effectiveness of Fusion loss} For solving the problem of cross-view, we devise a fusion loss strategy. And we set up two groups of experiments to verify the effectiveness of fusion loss by comparing the performances of single loss functions and the fusion loss function.  In each group of experiments, the pooling method was the same. \{row2, row3, row6\} was a group, they used pooling method JRPM(1 Scale), that was, global pooling, \{row4, row5, row16\} was another group, they used pooling method JRPM(3 Scales). From the experimental results, the fusion loss function significantly improved the accuracy of gait recognition than single triplet loss or arcface loss. The reason for this is that due to the skeleton data of the same person varies greatly from different viewpoints, although triplet loss do well in enlarging the inter-class distance and reducing the intra-class distance to a certain extent, the cross-view problem is still difficult for it. Besides, arcface loss can increase inter-class distance, may also increase intra-class distance.

\textbf{$\lambda $ setting} Previous analysis shows that the fusion of triplet loss and arcface loss need to be weighed. Because different viewpoints of a person may be very large, it may be difficult to find the cluster center, and triplet loss is good at dealing with it. Therefore, we consider that the triplet loss should play a major role, and $\lambda $ should be close to 1, and the results of experiments in \{row11, … , row18, row19\} have demonstrated it. With the increasing of $\lambda $ from 0.1 to 0.9, the recognition accuracy increased gradually, and then began to decline. In order to find a more accurate value, we purposely carried out experiments near $\lambda \text{=}0.9$, and the highest accuracy was $\lambda \text{=}0.9$ in the experiments, so the final setting was $\lambda \text{=}0.9$.
\begin{table}[htbp]
  \centering
  \caption{Ablation Experiments on CASIA-B}
     \scalebox{0.6} {
\begin{tabular}{r|r|r|r|rr}
      
    \midrule
    \multicolumn{2}{c|}{\multirow{2}[4]{*}{Strategies }} & \multicolumn{3}{c}{Accuracy (\%)} &  \\
\cmidrule{3-5}    \multicolumn{2}{c|}{} & \multicolumn{1}{c|}{NM} & \multicolumn{1}{c|}{BG} & \multicolumn{1}{c}{CL} &  \\
\cmidrule{1-5}    1     & Baseline(ST-GCN \cite{yan2018spatial}) & \multicolumn{1}{c|}{33.8} & \multicolumn{1}{c|}{23.6} & \multicolumn{1}{c}{12.5} &  \\
\cmidrule{1-5}    2     & triplet   + JRPM(S=1) & 43.8  & 27.8  & 22.7  &  \\
    3     & arcface + JRPM(S=1) & 43.2  & 29.2  & 13.7  &  \\
    4     & triplet   + JRPM(S=3) & 49.4  & 32.3  & 25.7  &  \\
    5     & arcface + JRPM(S=3) & 51.4  & 37.1  & 19.6  &  \\
    6     & 0.9 triplet + 0.1 arcface + JRPM(S=1) & 53.8  & 43.2  & 35.2  &  \\
    7     & 0.9 triplet + 0.1 arcface + JRPM(S=2) & 68.5  & 53.3  & 43.2  &  \\
    8     & 0.9 triplet + 0.1 arcface + JRPM(S=4) & 71.3  & 59.2 & 44.6  &  \\
    9     & 0.9 triplet + 0.1 arcface + JRPM(S=5) & 71.7  & \textbf{60.4}  & 45.9  &  \\
    10    & 0.9 triplet + 0.1 arcface + JRPM(S=6) & 70.2  & 55.8  & 43.7  &  \\
    11    & 0.1 triplet + 0.9 arcface + JRPM(S=3) & 59.0    & 50.8  & 41.5  &  \\
    12    & 0.5 triplet + 0.5 arcface + JRPM(S=3) & 63.8  & 49.8  & 39.7  &  \\
    13    & 0.8 triplet + 0.2 arcface + JRPM(S=3) & 70.5  & 57.4  & 41.1  &  \\
    14    & 0.85 triplet + 0.15 arcface + JRPM(S=3) & 71.6  & 58.1  & 43.6  &  \\
    15    & 0.875 triplet + 0.125 arcface + JRPM(S=3) & 72.1  & 57.7  & 43.9  &  \\
    16    & \textbf{0.9 triplet + 0.1 arcface + JRPM(S=3)} & \textbf{74.4} & 59.1  & \textbf{49.8} &  \\
    17    & 0.925 triplet + 0.075 arcface + JRPM(S=3) & 72.5  & 57.9  & 43.8  &  \\
    18    & 0.95 triplet + 0.05 arcface + JRPM(S=3) & 68.9  & 55.8  & 43.3  &  \\
    19    & 0.99 triplet + 0.01 arcface + JRPM(S=3) & 68.3  & 49.9  & 39.8  &  \\
\cmidrule{1-5}    
\end{tabular}}%
  \label{table1}%
\end{table}%

\textbf{Pyramid scales of JRPM} \{row6, … ,row10, row16\} shows the gait recognition accuracy of JRPM with different pyramid scales. We can judge from experiments results that JRPM has the satisfactory performance when pyramid scales are 3. With the number of pyramid scales increasing from 1 to 3, the recognition accuracy significantly improved from 53.8\% to 74.4\% in NM, from 43.2\% to 59.1 \% in BG, from 35.2\% to 49.8\% in CL, respectively. It showed that pyramid structure could enhance the discriminative ability of gait’s details. However, there is no significant improvement in 4 scales, 5 scales and 6 scales. Because more pyramid scales may add redundant information for gait recognition, besides bring additional computational cost. Therefore, 3 pyramid scales are employed in this work.

\subsection{Comparisons with model-based algorithms on Kinect Gait Biometry Dataset}

Table \ref{table2} reports the accuracy of our approach along with state-of-the-art model-based algorithms \cite{andersson2015person,yang2016relative,li2017dynamic,khamsemanan2017human,sun2018view,liu20193d}under experimental settings \cite{li2017dynamic} on Kinect Gait Biometry Dataset. As shown in Table \ref{table2}, the accuracy of JointsGait is 97.8\%, which is higher than the state-of-the-art approaches based manual feature\cite{andersson2015person,yang2016relative,khamsemanan2017human,sun2018view}, dynamic LSTM network\cite{li2017dynamic} or CNN-LSTM network\cite{liu20193d}. It is worth noting that JointsGait only uses corresponding 2D coordinates of joints, not 3D joints like other methods, which carry more information than 2D data. This proves the effectiveness of the proposed algorithm based on Gait GCNs and JRPM.

\begin{table}[htbp]
  \centering
  \caption{Comparisons with model-based algorithms on Kinect Gait Biometry Dataset}
    \scalebox{0.7} {
\begin{tabular}{c|c}
   
    \toprule
    Methods & Accuracy(\%) \\
    \midrule
    Andersson and Araujo \cite{andersson2015person}& 87.7 \\
    Yang et al. \cite{yang2016relative} & 94.5 \\
    Li et al.  \cite{li2017dynamic} & 96.6 \\
    Khamsemanan et al.\cite{khamsemanan2017human} & 97.0 \\
    Sun et al. \cite{sun2018view} & 82.7 \\
    Liu et al. \cite{liu20193d} & 97.4 \\
    \midrule
    JointsGait (ours) & \multirow{2}[2]{*}{\textbf{97.8}} \\
    (only using 16 joints with 2D coordinate) &  \\
    \bottomrule

    \end{tabular}}%
  \label{table2}%
\end{table}%

\begin{figure}[htb]
\centering
\includegraphics[scale=0.4]{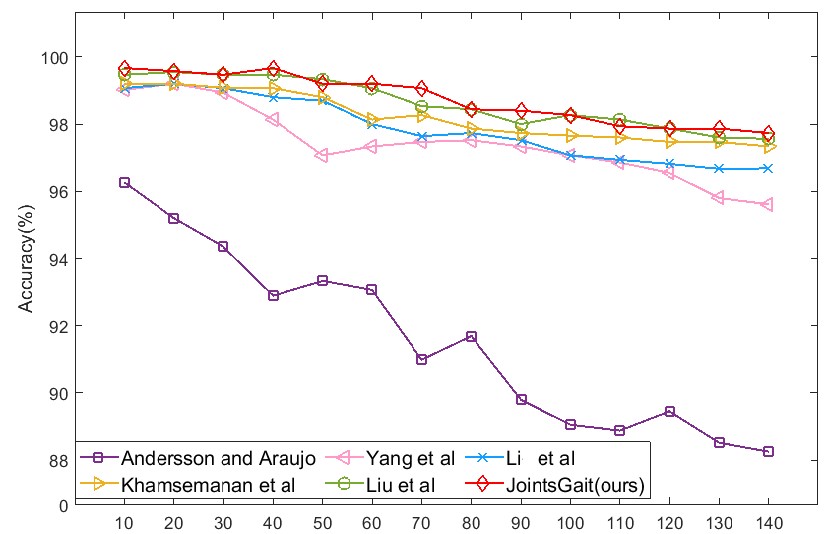}
\caption{ Performance comparison under different gallery sizes}
\label{fig7}
\end{figure}

Different gallery sizes may make gait recognition with different difficulty level. When the gallery size is small, the recognition accuracy may still be high even if a feature set’s performance is poor. Therefore, we further evaluated JointsGait with different Gallery sizes, in comparison to techniques from \cite{andersson2015person,yang2016relative,li2017dynamic,khamsemanan2017human,sun2018view,liu20193d}. For each size, individuals were randomly selected from the same 140 individuals\cite{li2017dynamic}. The results are shown in Figure\ref{fig7}. We can find that compared with other model-based methods, JointsGait is relatively less sensitive to changes in gallery size. The experimental results further validated that JointsGait has a sufficient capability to automatically learn the discriminative gait characteristics from 2D joints.

\subsection{Comparisons with model-based algorithms on CASIA-B Database}
We evaluate the robustness of JointsGait for cross-view and walking condition variations, and compare it with advanced model-based methods \cite{liao2017pose,an2018improving,liao2020model}  on CASIA-B Database. \cite{liao2017pose,an2018improving,liao2020model}  also use openpose \cite{cao2017realtime} to obtain 2D joints. But unlike us, PTSN-3D \cite{an2018improving}and PoseGait \cite{liao2020model} need additional calculations to estimate 3D joints from the 2D to deal with cross-view problem. 

Since there is no formal division of training and testing sets for this data set, we experiment with popular settings in the current literature \cite{liao2020model}, namely the train set consists of subjects 0 to 62 and the test set consists of subjects 63 to 124. In test sets, gallery maintains 1 to 4 clips of NM walking condition (NM \#1-4) , and probe retains 5 to 6 clips of NM walking condition(NM \#5-6), 1 to 2 clips of BG condition(BG \#1-2), and 1 to 2 clips of CL condition(CL \#1-2).
    
\begin{figure}[htb]
\centering
\includegraphics[scale=0.7]{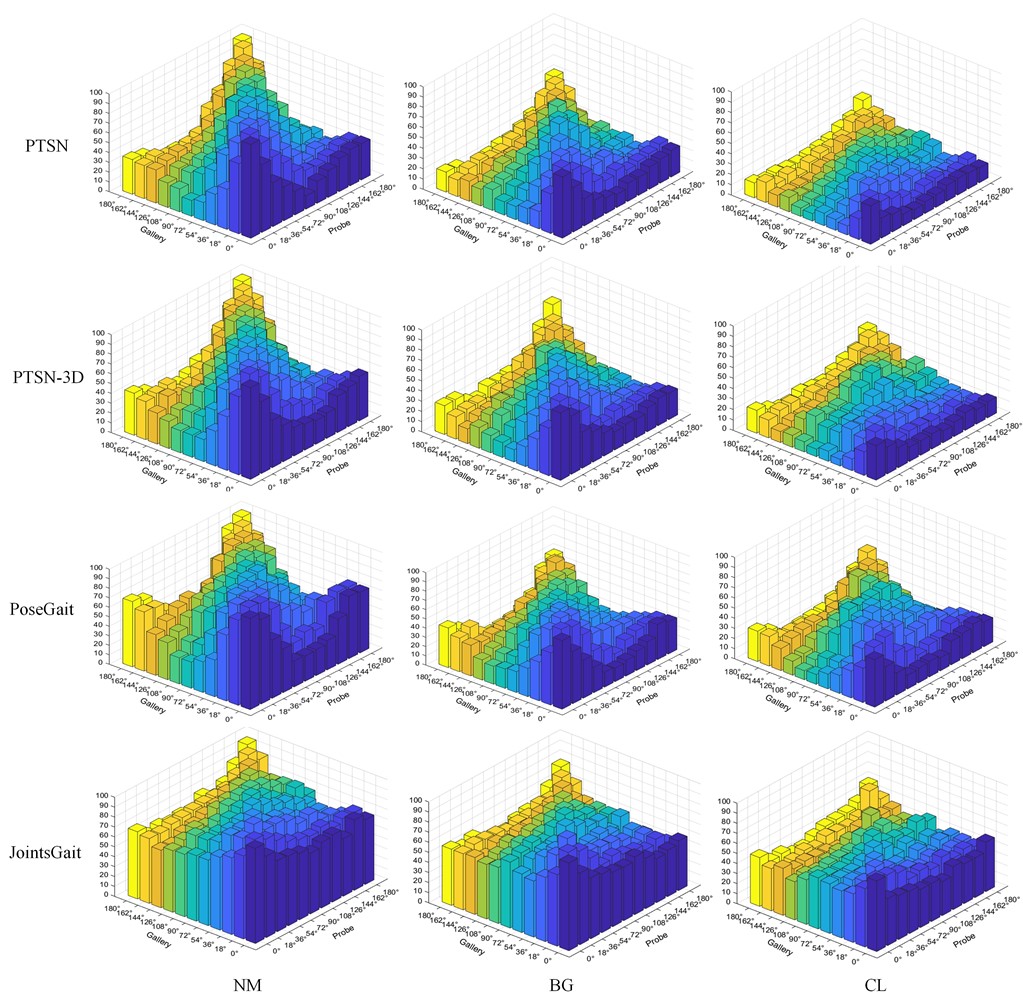}
\caption{Detailed recognition accuracies (\%) comparisons of the proposed method with state-of-art model-based algorithms.}
\label{fig8}
\end{figure}

The results are shown in Figure \ref{fig8} and Table \ref{table3}. Figure \ref{fig8} shows in detail the gait recognition accuracy of these model-based algorithms under cross-view of 11 views of gallery and 11 views of probe. It can be found in Figure \ref{fig8} that the recognition accuracy of other models fluctuate greatly, especially when the view difference reaches 90\degree, the recognition accuracy of other models are all very low, but the recognition accuracy of JointsGait in cross-view is not only high, but also stable in all three walking conditions. 

\begin{table}[htbp]
 \centering
  \caption{Average Recognition Rate (\%) Comparisons with model-based algorithms on CASIA-B Database , Excluding Identical-view Cases.}
 \scalebox{0.6} {
 \begin{tabular}{r|c|ccccccccccc|c|c}
\cmidrule{1-15}    \multicolumn{2}{c|}{Gallery} & \multicolumn{11}{c|}{0\degree-180\degree}                                                         & \multirow{2}[4]{*}{Average} & \multirow{2}[4]{*}{Std} \\
\cmidrule{1-15}    \multicolumn{2}{c|}{Probe} & 0\degree    & 18\degree   & 36\degree   & 54\degree   & 72\degree   & 90\degree   & 108\degree  & 126\degree  & 144\degree  & 162\degree  & 180\degree  &       &  \\
    \midrule
    \multicolumn{1}{c|}{NM} & PTSN \cite{liao2017pose} & 34.5  & 45.6  & 49.6  & 51.3  & 52.7  & 52.3  & 53    & 50.8  & 52.2  & 48.3  & 31.4  & 47.4  & 7.5 \\
    \multicolumn{1}{c|}{\#5-6} & PTSN-3D\cite{an2018improving} & 38.7  & 50.2  & 55.9  & 56    & 56.7  & 54.6  & 54.8  & 56    & 54.1  & 52.4  & 40.2  & 51.9  & 6.4 \\
          & PoseGait\cite{liao2020model} & 48.5  & 62.7  & 66.6  & 66.2  & 61.9  & 59.8  & 63.6  & 65.7  & 66    & 58    & 46.5  & 60.5  & 7 \\
         & JointsGait(ours) & \textbf{68.1} & \textbf{73.6} & \textbf{77.9} & \textbf{76.4} & \textbf{77.5} & \textbf{79.1} & \textbf{78.4} & \textbf{76} & \textbf{69.5} & \textbf{71.9} & \textbf{70.1} & \textbf{74.4} & \textbf{3.9} \\
    \midrule
    \multicolumn{1}{c|}{BG} & PTSN\cite{liao2017pose} & 22.4  & 29.8  & 29.6  & 29.2  & 32.5  & 31.5  & 32.1  & 31    & 27.3  & 28.1  & 18.2  & 28.3  & 4.4 \\
   \multicolumn{1}{c|}{\#1-2} & PTSN-3D\cite{an2018improving} & 27.7  & 32.7  & 37.4  & 35    & 37.1  & 37.5  & 37.7  & 36.9  & 33.8  & 31.8  & 27    & 34.1  & \textbf{3.9} \\
          & PoseGait\cite{liao2020model} & 29.1  & 39.8  & 46.5  & 46.8  & 42.7  & 42.2  & 42.7  & 42.2  & 42.3  & 35.2  & 26.7  & 39.6  & 6.6 \\
        & JointsGait (ours) & \textbf{54.3} & \textbf{59.1} & \textbf{60.6} & \textbf{59.7} & \textbf{63} & \textbf{65.7} & \textbf{62.4} & \textbf{59} & \textbf{58.1} & \textbf{58.6} & \textbf{50.1} & \textbf{59.1} & 4.2 \\
    \cmidrule{1-15}  \multicolumn{1}{c|}{CL} & PTSN\cite{liao2017pose} & 14.2  & 17.1  & 17.6  & 19.3  & 19.5  & 20    & 20.1  & 17.3  & 16.5  & 18.1  & 14    & 17.6  & \textbf{2.1} \\
    \multicolumn{1}{c|}{\#1-2} & PTSN-3D\cite{an2018improving} & 15.8  & 17.2  & 19.9  & 20    & 22.3  & 24.3  & 28.1  & 23.8  & 20.9  & 23    & 17    & 21.1  & 3.7 \\
          & PoseGait\cite{liao2020model} & 21.3  & 28.2  & 34.7  & 33.8  & 33.8  & 34.9  & 31    & 31    & 32.7  & 26.3  & 19.7  & 29.8  & 5.3 \\
          & JointsGait (ours) & \textbf{48.1} & \textbf{46.9} & \textbf{49.6} & \textbf{50.5} & \textbf{51} & \textbf{52.3} & \textbf{49} & \textbf{46} & \textbf{48.7} & \textbf{53.6} & \textbf{52} & \textbf{49.8} & 2.3 \\
    \bottomrule

    \end{tabular}
}
  \label{table3}%
\end{table}%
In Table \ref{table3}, results are the accuracies averaged on rest 10 views except the identical-view cases. For example, the accuracy of probe view 0\degree  is the average of 10 gallery views, and it does not include gallery view 0\degree. Then the average (Average) and the standard deviation (Std) of all these 11 average recognition accuracies are calculated. Std reflects the robustness of the model to the change of view angle. The lower the Std is, the more robust the model is to cross-view.

As illustrated in Table \ref{table3}, JointsGait not only achieves the best performance on average accuracy of 11 gallery views in all three walking conditions, respectively 74.4\% in NM, 59.1 \% in BG, 49.8 \% in CL, but also obtains highest recognition rates of each gallery view than other state-of-art model-based methods.

Similar to JointsGait, PTSN\cite{liao2017pose} use 2D joints obtained by openpose\cite{cao2017realtime} as input, but its average accuracy is lower than the proposed 27\% in NM, 30.8\% in BG, and 32.2\% in CL. Although PTSN-3D \cite{an2018improving} and PoseGait \cite{liao2020model} use 3D joints with more information, their performance are much worse than JointsGait, especially in carrying conditions and clothing changes. That means JointsGait can extract effective gait features in different walking conditions.

Besides, Although the 3D joints used by PTSN-3D \cite{an2018improving} and PoseGait \cite{liao2020model} are independent of views and can realize angle conversion, the average of Std in three walking conditions are higher than JointsGait (3.5) respectively 1.2 and 2.8. It is further proved that the proposed is robust to view changes. Therefore, the proposed method not only obtains high recognition accuracy but also can cope with the view variations.

\subsection{Comparisons with appearance-based algorithms on CASIA-B Database}
As mentioned before, due to the compactness of the model-based features, it is more difficult for model-based algorithms to extract feature than appearance-based ones. We compare JointsGait with advanced appearance-based gait recognition methods \cite{chao2019gaitset,yu2017invariant,yu2017gaitgan,he2018multi,yu2019gaitganv2,liu2020gait,zhang2020learning}to show the effectiveness of our model-based features when clothing variation seriously affect people’s appearance.

\begin{figure}[htb]
\centering
\includegraphics[scale=0.5]{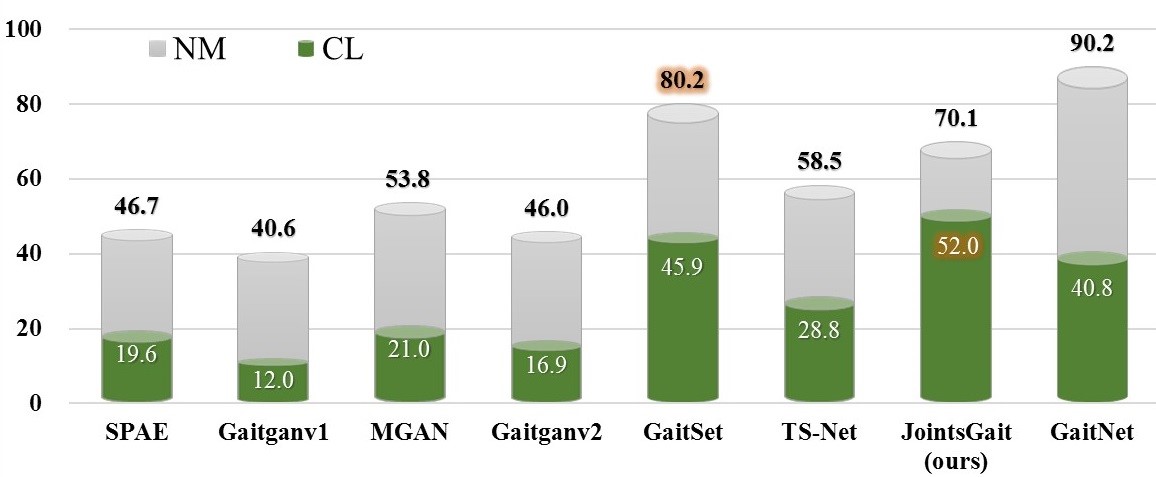}
\caption{The gap of Average recognition accuracy (\%) between NM and CL.}
\label{fig9}
\end{figure}

Compared with other views, when view is 180\degree, back view, clothing has the greatest influence on the person's appearance. Because style of clothes can be observed completely in back view, and there are more frames containing large-scale human shape in video frames. However, in practices back view is very common when pedestrians walk away from the surveillance camera. Hence, we evaluate JointsGait with appearance-based methods when the probe view is 180\degree. 
Specially, the train set and gallery set are same as Section 4.4, but probe retains 5 to 6 clips on 180\degree view of NM walking condition (NM \#5-6), and 1 to 2 clips 180\degree view of CL condition (CL \#1-2). Unlike othermethods, GaitNet\cite{zhang2020learning} is trained with all videos of the first 74 subjects and tested on the remaining 50 subjects. Results of these appearance-based algorithms are directly taken from their original paper in literature. Table \ref{table4}illustrates the comparison of the average recognition accuracy, and Figure \ref{fig9} is drawn according to Table \ref{table4} for intuitively showing the change of recognition accuracy of these methods from NM to CL.

\begin{table}[htbp]
  \centering
  \caption{Average Recognition Rate (\%) Comparisons with Appearance-based Methods when probe view is 180\degree in NM and CL, Excluding Identical-view Cases.}
    \scalebox{0.6} { \begin{tabular}{c|c|c|c|c}
    \toprule
    
          &       & \multicolumn{2}{c|}{Accuracy(\%)} & \multirow{3}[5]{*}{Rate of decline} \\
\cmidrule{3-4}          & \multirow{2}[3]{*}{Methods} & \multicolumn{2}{c|}{Probe 180\degree} &  \\
\cmidrule{3-4}          &       & NM  \#5-6 & CL \#1-2 &  \\
   \cmidrule{1-5}  \multirow{8}[3]{*}{Gallery NM\#1-4 (0\degree-162\degree)} & SPAE\cite{yu2017invariant} & 46.7  & 19.6  & 58.00\% \\
          & Gaitganv1\cite{yu2017gaitgan} & 40.6  & 12    & 70.40\% \\
          & MGAN\cite{he2018multi}& 53.8  & 21    & 61.00\% \\
          & Gaitganv2\cite{yu2019gaitganv2} & 46    & 16.9  & 63.30\% \\
          & GaitSet\cite{chao2019gaitset} & \textbf{80.2} & 45.9  & 42.80\% \\
          & TS-Net\cite{liu2020gait} & 58.5  & 28.8  & 50.80\% \\
          & JointsGait (ours) & 70.1  & \textbf{52} & \textbf{25.80\%} \\
\cmidrule{2-5}          & GaitNet\cite{zhang2020learning} & 90.2  & 40.8  & 54.80\% \\
    \bottomrule
   
    \end{tabular}}%
  \label{table4}%
\end{table}%

Although it is more challenging to compare with appearance-based algorithms, as shown in Table \ref{table4}, JointsGait achieves the highest recognition accuracy on CL even outperforms GaitNet\cite{zhang2020learning} with large train set, and is superior to most advanced appearance-based algorithms and is second only to GaitSet\cite{chao2019gaitset} in NM. This means that JointsGait is more robust and advantageous to the changes of clothing.

There are two reasons that our model-based method JointsGait is inferior to the appearance-based method GaitSet\cite{chao2019gaitset} in NM. One reason is GaitSet\cite{chao2019gaitset} uses high dimensional features, and the gait features we used consist of only body joints.  Another reason is that GaitSet\cite{chao2019gaitset} regards gait as an silhouette image set to enlarge the volume of training data while we use video sequence as input. Therefore, the volume of training data between us varies a lot. 

Although GaitSet\cite{chao2019gaitset} has achieved a very high performance in NM, its performance decrease severely in CL, about 42.8\%. As can be clearly seen from Figure \ref{fig9}, this case happens in all appearance-based methods. Even if GaitNet\cite{zhang2020learning} with large train set achieves 90.2 \% recognition accuracy in NM, its recognition accuracy on CL is only 40.8\%, whose rate of decline is over 50\%. Although Gaitganv1 \cite{yu2017gaitgan} has a recognition accuracy is only 40.6\% on NM, its recognition accuracy on CL decreases by as much as 70.4\%. However, the average recognition accuracy of JointsGait decreases only 25.8\% from NM to CL, besides JointsGait achieves satisfactory performance in NM. This is due to the features used in appearance-based algorithms often change greatly due to clothing variation. While JointsGait is robust to changes in walking conditions because it is based on the structure and motions of human body and is relatively insensitive to the appearance of human.

\section {Conclusions}
In this paper, we presented a novel model-based gait recognition method called JointsGait, whose input is 2D body joints. With the help of Joints Relationship Pyramid Mapping (JRPM) proposed by us, JointsGait can well learn local gait features from spatio-temporal gait features extracted by Gait GCNs. And a fusion loss strategy is designed to help JointsGait to be insensitive to cross-view and different walking conditions. 

Although JointsGait only uses 2D body joints, it surpasses those advanced model-based algorithms that use 3D body joints on the Kinect Gait Biometry Dataset. On CASIA-B database, the recognition accuracy of JointsGait greatly surpasses advanced model-based algorithms, and it performs well on cross-view problems. Besides, JointsGait outperforms state-of-art appearance-based algorithms when clothing seriously affects the appearance of people. The experimental results validated that JointsGait has a sufficient capability to automatically learn the discriminative gait characteristics from 2D joints. Hence, we can conclude that model-based gait recognition algorithms have great potential. 

However, 2D body joints are low dimensional features, which contain much less information than silhouettes used by appearance-based algorithms. How to retain the advantage of gait features extracted from joints and fully extract more features that are not affected by appearance will be the focus of our future work. 

\section*{Acknowledgments}
This work was supported by the National Natural Science Foundation of China [grant numbers 61231016 and 61871326].


\bibliography{mybibfile}

\begin{thebibliography}{10}
\expandafter\ifx\csname url\endcsname\relax
  \def\url#1{\texttt{#1}}\fi
\expandafter\ifx\csname urlprefix\endcsname\relax\def\urlprefix{URL }\fi
\expandafter\ifx\csname href\endcsname\relax
  \def\href#1#2{#2} \def\path#1{#1}\fi

\bibitem{nixon1999automatic}
M.~S. Nixon, J.~N. Carter, J.~M. Nash, P.~S. Huang, D.~Cunado, S.~V. Stevenage,
  Automatic gait recognition (1999) 13–18.

\bibitem{tanawongsuwan2001gait}
R.~Tanawongsuwan, A.~Bobick, Gait recognition from time-normalized joint-angle
  trajectories in the walking plane, in: Proceedings of the 2001 IEEE Computer
  Society Conference on Computer Vision and Pattern Recognition. CVPR 2001,
  Vol.~2, IEEE, 2001, pp. 726--731.

\bibitem{wang2004fusion}
L.~Wang, H.~Ning, T.~Tan, W.~Hu, Fusion of static and dynamic body biometrics
  for gait recognition, IEEE Transactions on circuits and systems for video
  technology 14~(2) (2004) 149--158.

\bibitem{han2005individual}
J.~Han, B.~Bhanu, Individual recognition using gait energy image, IEEE
  transactions on pattern analysis and machine intelligence 28~(2) (2005)
  316--322.

\bibitem{chao2019gaitset}
H.~Chao, Y.~He, J.~Zhang, J.~Feng, Gaitset: Regarding gait as a set for
  cross-view gait recognition, in: Proceedings of the AAAI Conference on
  Artificial Intelligence, Vol.~33, 2019, pp. 8126--8133.

\bibitem{cao2017realtime}
Z.~Cao, T.~Simon, S.-E. Wei, Y.~Sheikh, Realtime multi-person 2d pose
  estimation using part affinity fields, in: Proceedings of the IEEE conference
  on computer vision and pattern recognition, 2017, pp. 7291--7299.

\bibitem{wang2011human}
C.~Wang, J.~Zhang, L.~Wang, J.~Pu, X.~Yuan, Human identification using temporal
  information preserving gait template, IEEE transactions on pattern analysis
  and machine intelligence 34~(11) (2011) 2164--2176.

\bibitem{makihara2006gait}
Y.~Makihara, R.~Sagawa, Y.~Mukaigawa, T.~Echigo, Y.~Yagi, Gait recognition
  using a view transformation model in the frequency domain, in: European
  Conference on Computer Vision, Springer, 2006, pp. 151--163.

\bibitem{yu2017invariant}
S.~Yu, H.~Chen, Q.~Wang, L.~Shen, Y.~Huang, Invariant feature extraction for
  gait recognition using only one uniform model, Neurocomputing 239 (2017)
  81--93.

\bibitem{hu2013view}
M.~Hu, Y.~Wang, Z.~Zhang, J.~J. Little, D.~Huang, View-invariant discriminative
  projection for multi-view gait-based human identification, IEEE Transactions
  on Information Forensics and Security 8~(12) (2013) 2034--2045.

\bibitem{yu2017gaitgan}
S.~Yu, H.~Chen, E.~B. Garcia~Reyes, N.~Poh, Gaitgan: Invariant gait feature
  extraction using generative adversarial networks, in: Proceedings of the IEEE
  Conference on Computer Vision and Pattern Recognition Workshops, 2017, pp.
  30--37.

\bibitem{wu2016comprehensive}
Z.~Wu, Y.~Huang, L.~Wang, X.~Wang, T.~Tan, A comprehensive study on cross-view
  gait based human identification with deep cnns, IEEE transactions on pattern
  analysis and machine intelligence 39~(2) (2016) 209--226.

\bibitem{he2018multi}
Y.~He, J.~Zhang, H.~Shan, L.~Wang, Multi-task gans for view-specific feature
  learning in gait recognition, IEEE Transactions on Information Forensics and
  Security 14~(1) (2018) 102--113.

\bibitem{yu2019gaitganv2}
S.~Yu, R.~Liao, W.~An, H.~Chen, E.~B, Y.~Huang, N.~Poh, Gaitganv2: invariant
  gait feature extraction using generative adversarial networks, Pattern
  Recognition (2019) 179--189.

\bibitem{kastaniotis2016pose}
D.~Kastaniotis, I.~Theodorakopoulos, S.~Fotopoulos, Pose-based gait recognition
  with local gradient descriptors and hierarchically aggregated residuals,
  Journal of Electronic Imaging 25~(6) (2016) 063019.

\bibitem{andersson2015person}
V.~O. Andersson, R.~M. Araujo, Person identification using anthropometric and
  gait data from kinect sensor, in: Twenty-Ninth AAAI Conference on Artificial
  Intelligence, 2015, pp. 425--43.

\bibitem{yang2016relative}
K.~Yang, Y.~Dou, S.~Lv, F.~Zhang, Q.~Lv, Relative distance features for gait
  recognition with kinect, Journal of Visual Communication and Image
  Representation 39 (2016) 209--217.

\bibitem{li2017dynamic}
J.~Li, L.~Qi, A.~Zhao, X.~Chen, J.~Dong, Dynamic long short-term memory network
  for skeleton-based gait recognition, in: 2017 IEEE SmartWorld, Ubiquitous
  Intelligence \& Computing, Advanced \& Trusted Computed, Scalable Computing
  \& Communications, Cloud \& Big Data Computing, Internet of People and Smart
  City Innovation (SmartWorld/SCALCOM/UIC/ATC/CBDCom/IOP/SCI), IEEE, 2017, pp.
  1--6.

\bibitem{khamsemanan2017human}
N.~Khamsemanan, C.~Nattee, N.~Jianwattanapaisarn, Human identification from
  freestyle walks using posture-based gait feature, IEEE Transactions on
  Information Forensics and Security 13~(1) (2017) 119--128.

\bibitem{sun2018view}
J.~Sun, Y.~Wang, J.~Li, W.~Wan, D.~Cheng, H.~Zhang, View-invariant gait
  recognition based on kinect skeleton feature, Multimedia Tools and
  Applications 77~(19) (2018) 24909--24935.

\bibitem{liu20193d}
Y.~Liu, X.~Jiang, T.~Sun, K.~Xu, 3d gait recognition based on a cnn-lstm
  network with the fusion of skegei and da features, in: 2019 16th IEEE
  International Conference on Advanced Video and Signal Based Surveillance
  (AVSS), IEEE, 2019, pp. 1--8.

\bibitem{liao2017pose}
R.~Liao, C.~Cao, E.~B. Garcia, S.~Yu, Y.~Huang, Pose-based temporal-spatial
  network (ptsn) for gait recognition with carrying and clothing variations,
  in: Chinese Conference on Biometric Recognition, Springer, 2017, pp.
  474--483.

\bibitem{an2018improving}
W.~An, R.~Liao, S.~Yu, Y.~Huang, P.~C. Yuen, Improving gait recognition with 3d
  pose estimation, in: Chinese Conference on Biometric Recognition, Springer,
  2018, pp. 137--147.

\bibitem{liao2020model}
R.~Liao, S.~Yu, W.~An, Y.~Huang, A model-based gait recognition method with
  body pose and human prior knowledge, Pattern Recognition 98 (2020) 107069.

\bibitem{defferrard2016convolutional}
M.~Defferrard, X.~Bresson, P.~Vandergheynst, Convolutional neural networks on
  graphs with fast localized spectral filtering, in: Advances in neural
  information processing systems, 2016, pp. 3844--3852.

\bibitem{2015Gated}
Y.~Li, D.~Tarlow, M.~Brockschmidt, R.~Zemel, Gated graph sequence neural
  networks, Computer ence (2015).

\bibitem{hamilton2017inductive}
W.~Hamilton, Z.~Ying, J.~Leskovec, Inductive representation learning on large
  graphs, in: Advances in neural information processing systems, 2017, pp.
  1024--1034.

\bibitem{monti2017geometric}
F.~Monti, D.~Boscaini, J.~Masci, E.~Rodola, J.~Svoboda, M.~M. Bronstein,
  Geometric deep learning on graphs and manifolds using mixture model cnns, in:
  Proceedings of the IEEE Conference on Computer Vision and Pattern
  Recognition, 2017, pp. 5115--5124.

\bibitem{kipf2018neural}
T.~Kipf, E.~Fetaya, K.-C. Wang, M.~Welling, R.~Zemel, Neural relational
  inference for interacting systems, arXiv preprint arXiv:1802.04687 (2018).

\bibitem{he2015spatial}
K.~He, X.~Zhang, S.~Ren, J.~Sun, Spatial pyramid pooling in deep convolutional
  networks for visual recognition, IEEE transactions on pattern analysis and
  machine intelligence 37~(9) (2015) 1904--1916.

\bibitem{zhao2017pyramid}
H.~Zhao, J.~Shi, X.~Qi, X.~Wang, J.~Jia, Pyramid scene parsing network, in:
  Proceedings of the IEEE conference on computer vision and pattern
  recognition, 2017, pp. 2881--2890.

\bibitem{fu2019horizontal}
Y.~Fu, Y.~Wei, Y.~Zhou, H.~Shi, G.~Huang, X.~Wang, Z.~Yao, T.~Huang, Horizontal
  pyramid matching for person re-identification, in: Proceedings of the AAAI
  Conference on Artificial Intelligence, Vol.~33, 2019, pp. 8295--8302.

\bibitem{yan2018spatial}
S.~Yan, Y.~Xiong, D.~Lin, Spatial temporal graph convolutional networks for
  skeleton-based action recognition, arXiv preprint arXiv:1801.07455 (2018).

\bibitem{schroff2015facenet}
F.~Schroff, D.~Kalenichenko, J.~Philbin, Facenet: A unified embedding for face
  recognition and clustering, in: Proceedings of the IEEE conference on
  computer vision and pattern recognition, 2015, pp. 815--823.

\bibitem{hermans2017defense}
A.~Hermans, L.~Beyer, B.~Leibe, In defense of the triplet loss for person
  re-identification, arXiv preprint arXiv:1703.07737 (2017).

\bibitem{deng2019arcface}
J.~Deng, J.~Guo, N.~Xue, S.~Zafeiriou, Arcface\: Additive angular margin loss
  for deep face recognition, in: Proceedings of the IEEE Conference on Computer
  Vision and Pattern Recognition, 2019, pp. 4690--4699.

\bibitem{yu2006framework}
S.~Yu, D.~Tan, T.~Tan, A framework for evaluating the effect of view angle,
  clothing and carrying condition on gait recognition, in: 18th International
  Conference on Pattern Recognition (ICPR'06), Vol.~4, IEEE, 2006, pp.
  441--444.

\bibitem{liu2020gait}
X.~Liu, J.~Liu, Gait recognition method of underground coal mine personnel
  based on densely connected convolution network and stacked convolutional
  autoencoder, Entropy 22~(6) (2020) 695.

\bibitem{zhang2020learning}
Z.~Zhang, L.~Tran, F.~Liu, X.~Liu, On learning disentangled representations for
  gait recognition, IEEE Transactions on Pattern Analysis and Machine
  Intelligence (2020).

\end{thebibliography}

\end{document}